\documentclass[10pt,twocolumn,letterpaper]{article}

\usepackage{cvpr}
\usepackage{times}
\usepackage{epsfig}
\usepackage{graphicx}
\usepackage{amsmath}
\usepackage{amssymb}
\usepackage{subfigure}
\usepackage{float}
\usepackage{verbatim}
\usepackage{fancyhdr}
\usepackage{tabularx}
\usepackage{multirow}
\usepackage{booktabs}
\usepackage{makecell}
\usepackage[pagebackref=true,breaklinks=true,colorlinks,bookmarks=false]{hyperref}

\cvprfinalcopy % *** Uncomment this line for the final submission

\def\cvprPaperID{18} % *** Enter the CVPR Paper ID here

\ifcvprfinal\fi
\begin{document}

%%%%%%%%% TITLE
\title{Covariance Pooling for Facial Expression	Recognition}

\author{Dinesh Acharya$^\dagger$, Zhiwu Huang$^\dagger$, Danda Pani Paudel$^\dagger$, Luc Van Gool$^{\dagger\ddagger}$\\
	$^\dagger$Computer Vision Lab, ETH Zurich, Switzerland \quad $^\ddagger$VISICS, KU Leuven, Belgium\\
{\tt\small \{acharyad, zhiwu.huang, paudel, vangool\}@vision.ee.ethz.ch}
}

\maketitle
%\thispagestyle{empty}

%%%%%%%%% ABSTRACT
\begin{abstract}

% Classifying facial expressions into different categories requires capturing
% regional distortions of facial landmarks. We believe that second-order statistics such as covariance is better able to capture such distortions in regional facial features.
% In this work, we explore the benefits of using Riemannian networks for facial expression recognition. In particular, we employ Riemannian networks in conjunction with traditional convolutional neural networks for covariance pooling. By doing so, we were able to achieve total accuracy of $58.14\%$ on the
% validation set of Static Facial Expressions in the Wild (SFEW 2.0) and $87.0\%$
% on the validation set of Real-World Affective Faces (RAF) Database \footnote{Code will be eventually released on \url{https://github.com/d-acharya/CovPoolFER}}. Both of these results are the best results we are aware of. Covariance pooling can also be used to capture the temporal evolution of
% per-frame features. As an additional contribution, we also present results on advantage of using Riemanian Network to pool temporal image-set features.

Classifying facial expressions into different categories requires capturing
regional distortions of facial landmarks. We believe that second-order statistics such as covariance is better able to capture such distortions in regional facial features. 
In this work, we explore the benefits of using a manifold network structure for covariance pooling to improve facial expression recognition. In particular, we first employ such kind of manifold networks in conjunction with traditional convolutional networks for spatial pooling within individual image feature maps in an end-to-end deep learning manner. By doing so, we are able to achieve a recognition accuracy of $58.14\%$ on the
validation set of Static Facial Expressions in the Wild (SFEW 2.0) and $87.0\%$
on the validation set of Real-World Affective Faces (RAF) Database\footnote{The code of this paper will be eventually released on \url{https://github.com/d-acharya/CovPoolFER}}. Both of these results are the best results we are aware of. Besides, we leverage covariance pooling to capture the temporal evolution of
per-frame features for video-based facial expression recognition. Our reported results demonstrate the advantage of pooling image-set features temporally by stacking the designed manifold network of covariance pooling on top of convolutional network layers.

\end{abstract}

%%%%%%%%% BODY TEXT
\section{Introduction}
Facial expressions play an important role in communicating the state of our mind. Both humans and computer algorithms can greatly benefit from being able to classify facial expressions. Possible applications of automatic facial expression recognition include better transcription of videos, movie or advertisement recommendations, detection of pain in telemedicine etc.

\begin{figure}
		\begin{center}
				\includegraphics[width=.95\linewidth,keepaspectratio]{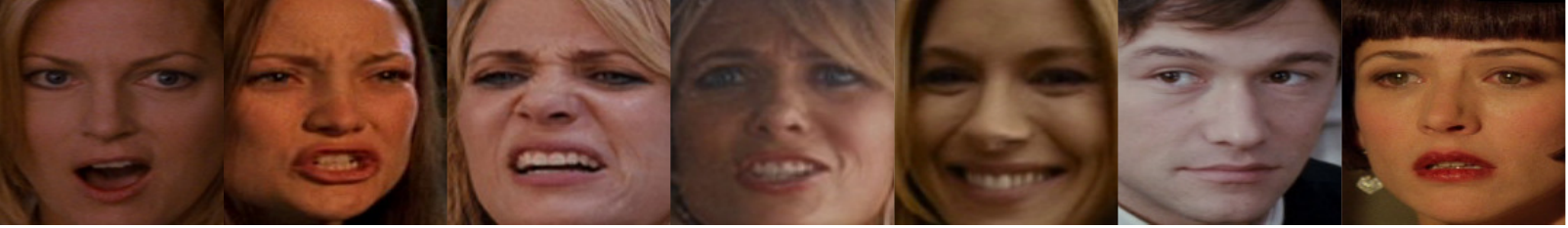}
				
					\includegraphics[width=.95\linewidth,keepaspectratio]{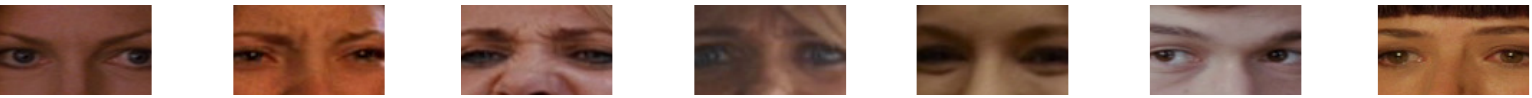}

		\end{center}
		\caption{Top: sample images of different facial expression classes from the SFEW dataset. Bottom: distortion of region between two eyebrows in the corresponding facial images.}
		\label{fig:sfew}
	\end{figure}

Traditional convolutional neural networks (CNNs) that use convolutional layers, max or average pooling and fully connected layers are considered to capture only first-order statistics \cite{epfl}. Second-order statistics such as covariance are considered to be better regional descriptors than first-order statistics such as mean or maximum \cite{cdu}. As shown in Figure \ref{fig:sfew}, facial expression recognition is more directly related to how facial landmarks are distorted rather than presence or absence of specific landmarks. We believe that second-order statistics is more suited to capture such distortions than first-order statistics. To learn second-order information deeply, we introduce covariance pooling after final convolutional layers. For further dimensionality reduction we borrow the concepts from the manifold network \cite{spdnet} and train it together with conventional CNNs in an end-to-end fashion. It is important to point out this is not a first work to introduce second-order pooling to traditional CNNs. Covariance pooling was initially used in \cite{mbprop} for pooling covariance matrix from the outputs of CNNs. \cite{epfl} proposed an alternative to compute second-order statistics in the setting of CNNs. However, such two works do not use either dimensionality reduction layers or non-linear rectification layers for second-order statistics. In this paper, we present a strong motivation for exploring them in the context of facial expression recognition.
% 	\begin{figure}
% 		\begin{center}
% 				\includegraphics[width=.95\linewidth,keepaspectratio]{res/sfew2}
% 		\end{center}
% 		\caption{Sample images from the SFEW dataset from different facial expression classes}
% 		\label{fig:sfew}
% 	\end{figure}
	
% 	\begin{figure}
% 		\begin{center}
% 				\includegraphics[width=.95\linewidth,keepaspectratio]{res/sfew3}
% 		\end{center}
% 		\caption{Distortion of region between two eyebrows across images from different facial expression classes}
% 		\label{fig:sfew}
% 	\end{figure}

In addition to being better able to capture distortions in regional facial features, covariance pooling can also be used to capture temporal evolution of per-frame features. Covariance matrix has been employed before to summarize per-frame features \cite{emoti14}. In this work, we experiment with using manifold networks for pooling per-frame features.

In summary, the contribution of this paper is two-fold:
\begin{itemize}
    \item End-to-end pooling of second-order statistics for both videos and images in the context of facial expression recognition
    \item State-of-art result on image-based facial expression recognition

\end{itemize}

\begin{comment}
In following section, we briefly review relevant prior works for both image-based and video-based facial recognition problems. In section 2, we discuss in detail the datasets used for our work and alignment techniques used. In section 3, we discuss about covariance pooling. We further discuss about applying covariance pooling to pool temporal features. In section 5, we present and discuss experimental results. We end with concluding remarks and possible extensions.
\end{comment}

	\section{Related Works}
	Though facial expression recognition from both images and videos are closely related, they each have their own challenges. Videos contain dynamic information which a single image lacks. With this additional dynamic information, we should theoretically be able to improve facial expression accuracy. However, extracting information from videos has its own challenges. In following sub-sections, we briefly review standard approaches to facial expressions on both image and video-based approaches.
		
	\subsection{Facial Expression Recognition from Images}
	
	Most of the recent approaches in facial expression recognition from images use various standard architectures such as VGG networks, Inception networks, Residual networks, Inception-Residual Networks etc \cite{grimaces}\cite{face2exp}\cite{regularizefacenet}. Many of these works carry out pretraining on FER-2013, face recognition datasets or similar datasets and either use outputs from fully connected layers as features to train classifiers or fine-tune the whole network. Use of ensemble of multiple CNNs and fusion of the predicted scores is also widely used and found to be successful. For example, in Emotiw2015 sub challenge on image-based facial expression recognition, both winners and runner up teams \cite{committee}\cite{cmu} employed ensemble of CNNs to achieve the best reported score. There, pre-training was done on FER-2013 dataset. Recently, in \cite{grimaces}, authors reported validation accuracy of $54.82\%$ which is a state-of-art result for a single network. The accuracy was achieved using VGG-VD-16. The authors carried out pre-training on VGGFaces and FER-2013.
				
	All such networks discussed above employ traditional neural network layers. These architectures can be considered to capture only first-order statistics. Covariance pooling, on the other hand captures second-order statistics. One of the earliest works employing covariance pooling for feature extraction used it as regional descriptor \cite{semanticsegmentation}\cite{cdu}. In \cite{epfl}, authors propose various architectures based on VGG network to employ covariance pooling. In \cite{spdnet}, authors present a deep learning architecture for learning on Riemannian manifold which can be employed for covariance pooling.
		
	\subsection{Facial Expression Recognition from Videos}
	
	Traditionally, video-based recognition problems used per-frame features such as SIFT, dense-SIFT, HOG \cite{emoti14} and recently deep features extracted with CNNs have been used \cite{emoti161} \cite{emoti167}. The per-frame features are then used to assign score to each individual frame. Summary statistics of such per-frame features are then used for facial expression recognition. In \cite{emoti151}, authors propose modification of Inception architecture to capture action unit activation which can be beneficial for facial expression recognition.	Other works use various techniques to capture the temporal evolution of the per-features. For example, LSTMs have been successfully employed with various names such as CNN-RNN, CNN-BRNN etc \cite{emoti153}\cite{emoti161}\cite{emoti1632}. 3D convolutional neural networks have also been used for facial expression recognition. However, performance of a single 3D-ConvNet was worse than applying LSTMs on per-frame features \cite{emoti161}. State-of-art result reported in~\cite{emoti161} was obtained by score fusion of multiple models of 3D-ConvNets and CNN-RNNs.
		
	Covariance matrix representation was used as one of the summary statistics of per-frame features in \cite{emoti14}. Kernel-based partial least squares (PLS) were then used for recognition. Here, we use the methods in \cite{emoti14} as baseline and use the SPD Riemannian networks instead of kernel based PLS for recognition and obtain slight improvement.
%-------------------------------------------------------------------------

	\section{Facial Expression Recognition and Covariance Pooling}
	
% 	\subsection{Image-based Facial Expression Recognition}
% 	Facial expression is localized in the facial region whereas images in the wild contain large irrelevant information. Due to this, face detection is performed first and then aligned based on facial landmark locations. Various techniques such as pretrained convolutional neural networks or traditional hand crafted feature extraction techniques can then be applied to the normalized face. Classifiers are then used to classify such features.
	
% 	\subsection{Video-based Facial Expression Recognition}
%     As the case of image-based facial expression recognition, videos in the wild contain large irrelevant information. First, all the frames are extracted from a video. Face detection and alignment is then performed on each individual frame. Depending on the feature extraction algorithm, either image features are extracted from the normalized faces or the normalized faces are concatenated and 3d convolutions are applied to the concatenated frames.

\subsection{Overview}
Facial expression is localized in the facial region whereas images in the wild contain large irrelevant information. Due to this, face detection is performed first and then aligned based on facial landmark locations. Next, we feed the normalized faces into a deep CNN. To pool the feature maps spatially from the CNN, we propose to use covariance pooling, and then employ the manifold network \cite{spdnet} to deeply learn the second-order statistics. The pipeline of our proposed model for image-based facial expression recognition is shown in Figure \ref{fig:imageCovMatrix}.

As the case of image-based facial expression recognition, videos in the wild contain large irrelevant information. First, all the frames are extracted from a video. Face detection and alignment is then performed on each individual frame. Depending on the feature extraction algorithm, either image features are extracted from the normalized faces or the normalized faces are concatenated and 3d convolutions are applied to the concatenated frames. Intuitively, as the temporal convariance can capture the useful facial motion pattern, we propose to pool the frames over time. To deeply learn the temporal second-order information, we also employ the manifold network \cite{spdnet} for dimensionality reduction and non-linearity on covariance matrices. The overview of our presented model for video-based facial expression recognition is illustrated in Figure \ref{fig:videoCovMatrix}.
    
Accordingly, the core techniques of the two proposed models are spatial/temporal covariance pooling and the manifold network for learning the second-order features deeply. In the following we will introduce the two crucial techniques.

\subsection{Covariance Pooling}
As discussed earlier, traditional CNNs that consist of fully connected layers, max or average pooling and convolutional layers only capture first-order information \cite{epfl}. ReLU introduces non-linearity but does so only at individual pixel level. Covariance matrices computed from features are believed to be better able to capture regional features than first-order statistics \cite{cdu}.
		
Given a set of features, covariance matrix can be used to compactly summarize the second-order information in the set. If $\mathbf{f_1},\mathbf{f_2},\dots,\mathbf{f_n}\in\mathbb{R}^d$ be the set of features, the covariance matrix can be computed as:
	
		\begin{equation}
	    	\label{eqn:1}
		    \mathbf{C}=\frac{1}{n-1}\sum_{i=1}^n(\mathbf{f_i}-\mathbf{\bar{f}})(\mathbf{f_i}-\mathbf{\bar{f}})^T,
		\end{equation}
	where $\mathbf{\bar{f}}=\frac{1}{n}\sum_{i=1}^n\mathbf{f_i}$.

	The matrices thus obtained are symmetric positive definite (SPD) only if number of linearly independent components in $\{\mathbf{f_1},\mathbf{f_2},\dots,\mathbf{f_n}\}$ is greater than $d$. In order to employ the geometric structure preserving layers of the SPD manifold network \cite{spdnet}, the covariance matrices are required to be SPD. However, even if the matrices are only positive semi definite, they can be regularized by adding a multiple of trace to diagonal entries of the covariance matrix:
		\begin{equation}
		\label{eqn:2}
		    \mathbf{C^+}=\mathbf{C}+\lambda trace(\mathbf{C})\mathbf{I},
		\end{equation}
	where $\lambda$ is a regularization parameter and $\mathbf{I}$ is identity matrix.
	
% 	\subsubsection{Covariance Matrix from Convolutional Layers}
		\begin{figure}
		\begin{center}
			\includegraphics[width=.95\linewidth,keepaspectratio]{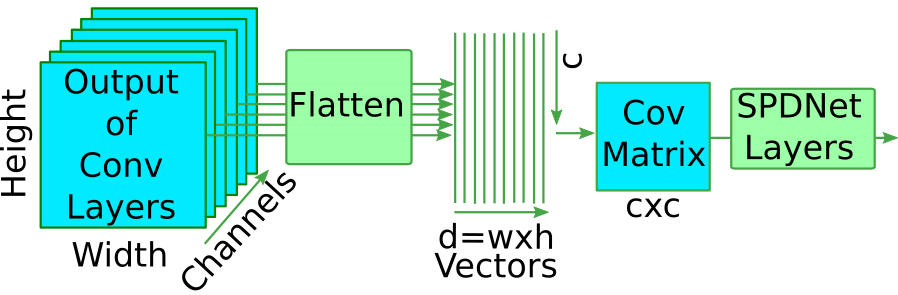}
		\end{center}
		\caption{In order to leverage covariance pooling on image-based facial expression recognition problem, output of convolutional layer is flattened as illustrated. The covariance matrix is computed form resulting vectors.}
		\label{fig:imageCovMatrix}
	\end{figure}
	
		\paragraph{Covariance Matrix for Spatial Pooling:}
		In order to apply covariance pooling to image-based facial expression recognition problem, as shown in Figure \ref{fig:imageCovMatrix}, outputs from final convolutional layers can be flattened and used to compute covariance matrix. Let $\mathbf{X} \in \mathbb{R}^{w\times h\times d}$ be the output obtained after several convolutional layers, where $w,h,d$ stand for width, height and number of channels in the output respectively. $\mathbf{X}$ can be flattened as an element $\mathbf{X'}\in\mathbb{R}^{n\times d}$ where $n=w\times h$. If $\mathbf{f_1},\mathbf{f_2},...,\mathbf{f_n}\in\mathbb{R}^d$ be columns of $\mathbf{X'}$, we can capture the variation across channels by computing covariance as in Eqn~\ref{eqn:1} and regularizing thus computed matrix using Eqn.~\ref{eqn:2}.
		
% 		\subsubsection{Covariance Matrix from Image Set Features}
		
	    \begin{figure}
		\begin{center}
		    \includegraphics[width=.95\linewidth,keepaspectratio]{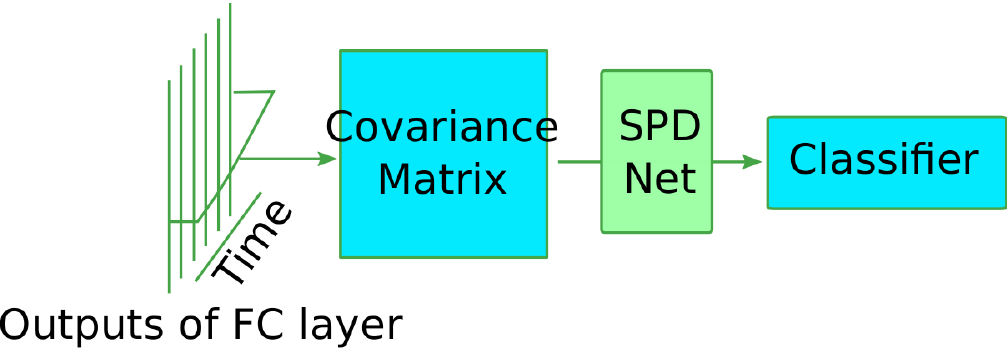}
		\end{center}
		\caption{In case of video-based facial expression recognition problems, output of fully connected layers are considered as image set features. The covariance matrix is computed from such image set features.}
		\label{fig:videoCovMatrix}
	\end{figure}
	
	\paragraph{Covariance Matrix for Temporal Pooling:}
		As illustrated in Figure \ref{fig:videoCovMatrix}, covariance pooling can be employed in \cite{emoti14} to pool temporal features. If $\mathbf{f_1},\mathbf{f_2},\dots,\mathbf{f_n}\in \mathbb{R}^d$ be per-frame features extracted from images, we can compute covariance matrix using the Eqn.~\ref{eqn:1} and regularize it using Eqn.~\ref{eqn:2}.

		\subsection{SPD Manifold Network (SPDNet) Layers}
		The covariance matrices thus obtained typically reside on the Riemannian manifold of SPD matrices. Directly flattening and applying fully connected layers directly causes loss of geometric information. Standard methods apply logarithm operation to flatten the Riemannian manifold structure to be able to apply standard loss functions of Euclidean space \cite{semanticsegmentation}\cite{cdu}. The covariance matrices thus obtained are often large and their dimension needs to be reduced without losing geometric structure. In \cite{spdnet}, authors introduce special layers for reducing dimension of SPD matrices and to flatten the Riemannian manifold to be able to apply standard loss functions.

		In this subsection, we briefly discuss the layers introduced in~\cite{spdnet} for learning on Riemannian Manifold.
		\paragraph{Bilinear Mapping Layer (BiMap)} 
		Covariance matrices computed from features can be large and it may not be feasible to directly apply fully connected layers after flattening them. Furthermore, it is also important to preserve geometric structure while reducing dimension. The BiMap layer accomplishes both of these conditions and plays the same role as traditional fully connected layers. If $\mathbf{X}_{k-1}$ be input SPD matrix, $\mathbf{W}_k\in \mathbb{R}_{*}^{d_k\times d_{k-1}}$ be weight matrix in the space of full rank matrices and $\mathbf{X}_k\in\mathbb{R}^{d_k\times d_k}$  be output matrix, then $k$-th the bilinear mapping $f_b^k$ is defined as 
		\begin{equation}
		    \mathbf{X}_k=f_b^{k}(\mathbf{X}_{k-1};\mathbf{W}_k)=\mathbf{W}_k\mathbf{X}_{k-1}\mathbf{W}_k^T.
		\end{equation}
		
		\paragraph{Eigenvalue Rectification (ReEig)}
		ReEig layer can be used to introduce non-linearity in the similar way as Rectified Linear Unit (ReLU) layers in traditional neural networks. If $\mathbf{X}_{k-1}$ be input SPD matrix, $\mathbf{X}_{k}$ be output and $\epsilon$ be eigenvalue rectification threshold, $k$-th ReEig Layer $f_r^{k}$ is defined as:
		\begin{equation}
		    \mathbf{X}_k=f_r^{k}(\mathbf{X}_{k-1})=\mathbf{U}_{k-1}\max(\epsilon\mathbf{I},\sigma_{k-1})\mathbf{U}_{k-1}^T,
		\end{equation}
		where $\mathbf{U}_{k-1}$ and $\mathbf{\Sigma}_{k-1}$ are defined by eigenvalue decomposition $\mathbf{X}_{k-1}=\mathbf{U}_{k-1}\Sigma_{k-1}\mathbf{U}_{k-1}^T$. The $\max$ operation is element-wise matrix operation.
		
		\paragraph{Log Eigenvalue Layer (LogEig)}
		As discussed earlier, SPD matrices lie on Riemannian manifold. The final LogEig layer endows elements in Riemannian manifold with a Lie Group structure so that matrices can be flattened and standard euclidean operations can be applied. If $\mathbf{X}_{k-1}$ be input matrix, $\mathbf{X}_k$ be output matrix, the LogEig layer applied in $k$-th layer $f_l^k$ is defined as 
		\begin{equation}
		    \mathbf{X}_k = f^{k}_l(\mathbf{X}_{k-1})=\log(\mathbf{X}_{k-1}) = \mathbf{U}_{k-1}\log(\Sigma_{k-1})\mathbf{U}_{k-1}^T,
		\end{equation}
		where $\mathbf{X}_k=\mathbf{U}_{k-1}\Sigma_{k-1}\mathbf{U}_{k-1}^T$ is an eigenvalue decomposition and $\log$ is an element-wise matrix operation.
		
\begin{figure}
		\begin{center}
				\includegraphics[width=.95\linewidth,keepaspectratio]{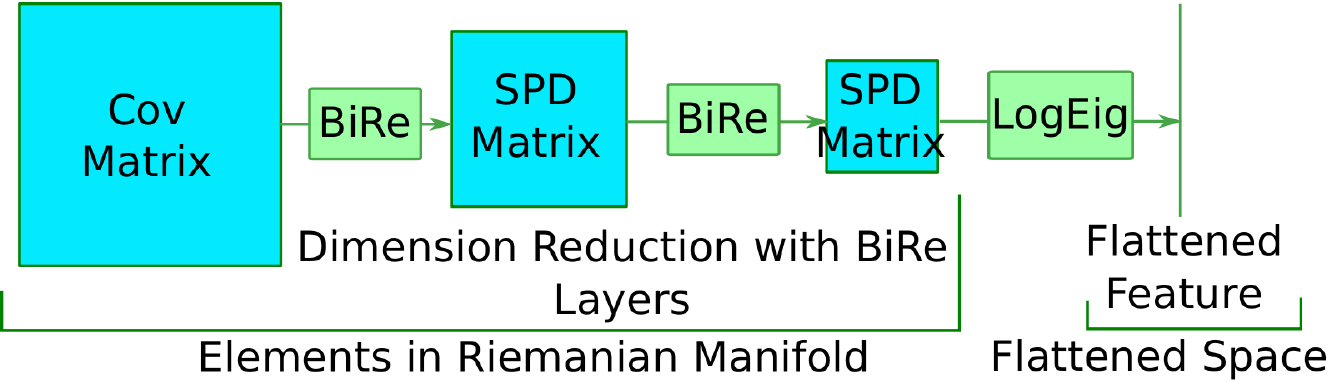}
		\end{center}
		\caption{Illustration of SPD Manifold Network (SPDNet) with 2-BiRe layers.}
		\label{fig:spdnetconcept}
	\end{figure}

		BiMap and ReEig layers can be used together as a block and is abbreviated as BiRe. The architecture of a SPDNet with 2-BiRe layers is shown in Figure \ref{fig:spdnetconcept}.

\section{Experiments}
	\subsection{Benchmark Datasets}
Datasets that contain samples with either real or acted facial expressions in the wild were chosen. Such datasets are better approximation to the real world scenarios than posed datasets and are also more challenging.
	\paragraph{Image-based Facial Expression Recognition}
	
	For comparing our deep learning architectures for image-based facial expression recognition against standard results, we use Static Facial Expressions in the Wild (SFEW) 2.0 \cite{sfew} \cite{afew} dataset and Real-world Affective Faces (RAF) dataset \cite{dlpcnn}. SFEW 2.0 contains 1394 images, of which 958 are to be used for training and 436 for validation. This dataset was prepared by selecting frames from videos of AFEW dataset. Facial landmark points provided by the authors were detected using mixture-of-parts based model \cite{zhuandramanan}. The landmarks thus obtained were then used for alignment. The RAF dataset \cite{dlpcnn} was prepared by collecting images from various search engines and the facial landmarks were annotated manually by 40 independent labelers. The dataset contains 15331 images labeled with seven basic emotion categories of which 3068 are to be used for validation and 12271 for training.
	
	It is worth pointing out that there exist several other image-based datasets such as EmotioNet~\cite{emotionet} and FER-2013~\cite{goodfellow}. However, they have their own downsides. Though EmotioNet is the largest existing dataset for facial expression recognition, the images were automatically annotated and the labels are incomplete. FER-2013, contains relatively small image size and does not contain RGB information. Most other databases either contain too few samples or are taken in well posed laboratory setting.
	
	\paragraph{Video-based Facial Expression Recognition}
	For video-based facial expression recognition, we use Acted Facial Expressions in the Wild (AFEW) dataset to compare our methods with existing methods. This dataset was prepared by selecting videos from movies. It contains about 1156 publicly available labeled videos of which 773 videos are used for training and 383 for validation. Just as in case of SFEW 2.0 dataset, the landmarks and aligned images provided by authors were obtained using mixture-of-parts based model.
	
	Though there exist several other facial expression recognition databases for videos such as Cohn-Kanade/Cohn-Kanade+ (CK/CK+) \cite{ck}\cite{ckp},  most of them are either sampled in well controlled laboratory environment or are labeled with action unit encoding rather than seven basic classes of facial expressions.
	
	\subsection{Face Detection and Alignment}
	Authors of RAF database \cite{dlpcnn} provide manually annotated face landmarks, while those of SFEW 2.0 \cite{sfew} and AFEW \cite{afew} datasets do not and instead provide landmarks and aligned images obtained using mixture-of-parts based model \cite{zhuandramanan}.
	%\cite{alignzhu}.
	Images and videos captured in the wild contain large amount of non-essential information. Face detection and alignment helps remove non-essential information from the data samples. Furthermore, to be able to compare variations in local facial features across images, face alignment is important. This serves as normalization of data. While trying to categorize facial expressions from videos, motion of people, change of background etc. also lead to large unwanted variation across image frames. Due to this, training algorithms on original unaligned data is not feasible. Face alignment additionally helps to capture the dynamic evolution of local facial features across images of the same videos in an effective manner.
	
	For face and facial landmark detection Multi-task Cascade Convolutional Neural Networks (MTCNN) \cite{mtcnn} was used. MTCNN was found to be more accurate and successful for alignment compared to mixture-of-parts based model. After successful face and facial landmark detection, we use three points constrained affine transformation for face alignment. Coordinates of left eye, right eye and midpoint of corners of the lips were used for alignment.

	\subsection{Baseline Model and Architectures for Image-based Problem}
	\paragraph{Comparison of Standard Architectures}
		In Table \ref{table:detectionrate2} we present the comparison of accuracies of training or finetuning various standard network architectures. For a baseline model, we take the network architecture presented in \cite{dlpcnn}. The scores reported on RAF database for VGG network and AlexNet in \cite{dlpcnn} is less compared to their base line model. So the networks are not trained again here. It is worth pointing out that there, authors report per class average accuracy but we report total accuracy only here.
		\begin{table}
			\centering
			\begin{tabular}{lcc}
				\toprule
				\multirow{2}{*}{Models} &
				RAF & SFEW 2.0\\
				&Total & Total\\
				\midrule
				VGG-VD-16 network\cite{grimaces} & - & 54.82\\
				\makecell{Inception-ResnetV1\\  (Trained from scratch)$^\ddagger$ }&   82.6 & 47.37 \\
				\makecell{Inception-ResnetV1 \\ (Finetuned) $^\ddagger$   }& 83.4 & 51.9\\ 
				\makecell{Baseline Model} $^\ddagger$   &  84.5  & 54.45\\
				\bottomrule
			\end{tabular}
			\caption{Comparison of image-base recognition accuracies of various standard models on validation set of the RAF and SFEW 2.0 datasets. Here the models labelled $^\ddagger$ were trained on our own.}
			\label{table:detectionrate2}
		\end{table}
	Here, we use center loss\cite{centerloss} to train the network in all cases rather than locality preserving loss\cite{dlpcnn} as we do not deal with compound emotions. In all cases, dataset was augmented using standard techniques such as random crop, random rotate and random flip. For SFEW 2.0, in all cases, output of second to last fully connected layer was used as image features and Support Vector Machines (SVMs) were trained separately. Note that the models labelled $^\ddagger$ were trained on our own. Inception-ResnetV1~\cite{inceptionnet} was both trained from scratch, as well as finetuned on a model pre-trained on subset of MS-Celeb-1M dataset. It is evident from the table that fine-tuning the Inception-ResnetV1 trained on face recognition dataset performs better than training from scratch. It should not come as a surprise that a relatively small network outperforms Inception-ResNet model as there are more parameters to be learned in deeper models. For further experiments and to introduce covariance pooling, we use the baseline model from \cite{dlpcnn}.
	\paragraph{Incorporation of SPD Manifold Network}
	As discussed above, we introduce covariance pooling and subsequently the layers from the SPD manifold network (SPDNet) after the final convolutional layer. While introducing covariance pooling, we experimented with various models for the architecture. The details of the various models considered are summarized in Table~\ref{table:architecture_list}.
	\begin{table}[ht]
		\centering
		\begin{tabular}{@{} p{1cm} p{1.1cm} p{1.25cm} p{1.25cm} p{1.25cm} @{}}
			\toprule
			Baseline & Mode-1 & Model-2 & Model-3 & Model-4\\
			\midrule
			Conv256 	& Conv256 	& Conv256 	& Conv256 	& Conv256\\
			\midrule
			 			& Cov 		& Cov 		& Cov		& Cov \\
			& BiRe 		& BiRe 		& BiRe		& BiRe \\
			& LogEig	& LogEig 	& BiRe		& LogEig \\
			&  			&  			& LogEig	& \\
			&  			&  			& 			& \\
			FC2000	& FC2000 	& FC2000	& FC2000	& FC2000\\
			FC7		& FC7 		& FC128 	& FC7		& FC512\\
			&  			& FC7		& 			& FC7\\
			\bottomrule
		\end{tabular}
		\caption{Various models considered for covariance pooling. For brevity, initial convolution layers are ignored.}
		\label{table:architecture_list}
	\end{table}	
	\subsection{Results on Image-based Problem}
    Covariance pooling was applied after final convolution layer and before fully connected layers. Various models described in Table \ref{table:architecture_list} and their accuracies are listed below in Table \ref{table:modacc}.
		\begin{table}[ht]
			\centering
			\begin{tabular}{lccc}
				\toprule
				\multirow{2}{*}{Model}	&  RAF	& SFEW 2.0\\ 
				& Total Accuracy	& Total Accuracy \\
				\midrule
				Baseline Model ~\cite{dlpcnn}  & 84.7  			& 54.45 \\ 
				Model-1 				& 86.3  			& 55.40 \\
				Model-2 				& \textbf{87.0} 				& 56.72 \\
				Model-3 				& 85.0 				& 57.48 \\ 
				Model-4 				& 85.4 				& \textbf{58.14}\\
				\midrule
				\midrule
				VGG-VD-16~\cite{grimaces}				&  -					& 54.82\\
				EmotiW-1 (2015)~\cite{cmu}& -				& 55.96\\
				EmotiW-2 (2015)~\cite{committee}		& -					& 52.80\\
				\bottomrule
			\end{tabular}
			\caption{Image-based recognition accuracies for various models with and without covariance pooling.}
			\label{table:modacc}
		\end{table}
		For the RAF database, as stated earlier, the network was trained in end-to-end fashion. However, for SFEW 2.0 dataset, we use output of penultimate fully connected layer (which ranges from 128 to 2000 dimensional feature depending on the model considered). It is worth pointing out that for SFEW 2.0 our single model performed better than ensemble of convolutional neural networks in \cite{cmu} and \cite{committee}. It could be argued that the datasets used for pre-training were different in our case and in \cite{cmu}\cite{committee}. However, improvement of almost $3.7\%$ over baseline in the SFEW 2.0 dataset justifies the use of SPDNet for facial expression recognition.
		
		It is also important to point out that on the SFEW 2.0 and AFEW datasets, face detection failed in several images and videos. To report validation score, we assign random uniform probability of success ($\frac{1}{7}$) for correct recognition to the samples on which face detection did not succeed.
		
				\begin{table*}[!ht]
		    \centering
		    \begin{tabular}{m{2cm}m{5cm}m{5cm}m{4cm}}
		    Original Class & Correctly Predicted & Incorrectly Predicted & Predicted Classes\\
		    Angry& \begin{minipage}{.3\textwidth}
		   \includegraphics[width=5cm]{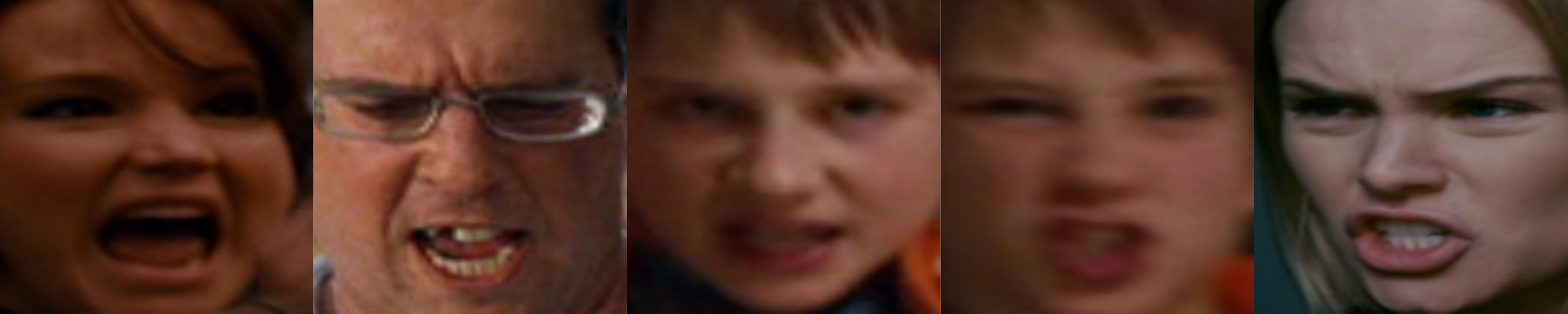} 
		   \end{minipage}
		   & \begin{minipage}{.3\textwidth}
		   \includegraphics[width=5cm]{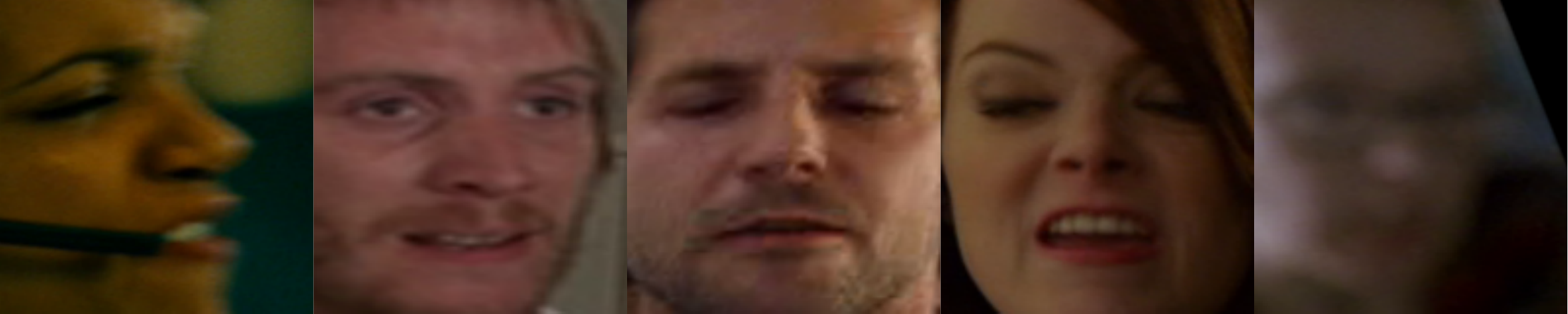}
		    \end{minipage} & Neutral, Neutral, Neutral, Neutral, Happy\\
		    
		     Disgust& \begin{minipage}{.3\textwidth}
		   \includegraphics[width=5cm]{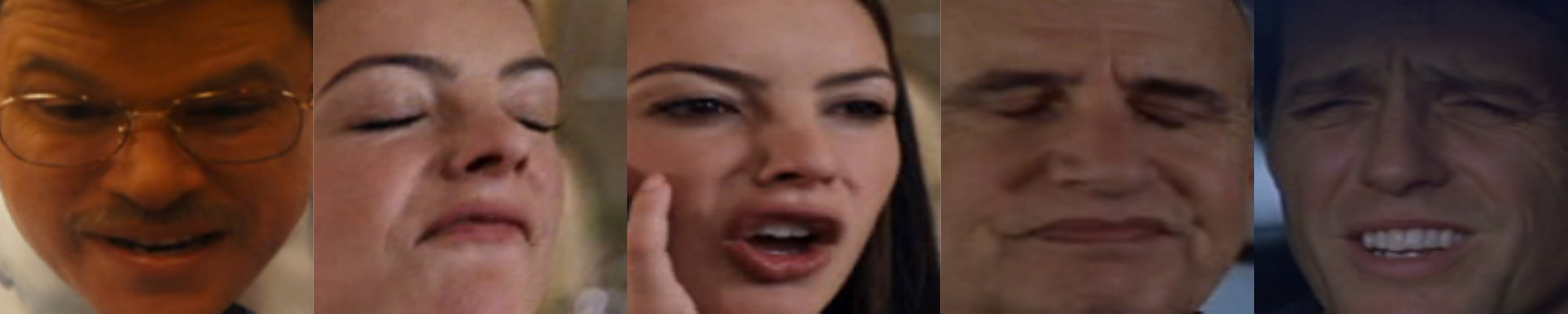} 
		   \end{minipage}
		   & \begin{minipage}{.3\textwidth}
		   \includegraphics[width=5cm]{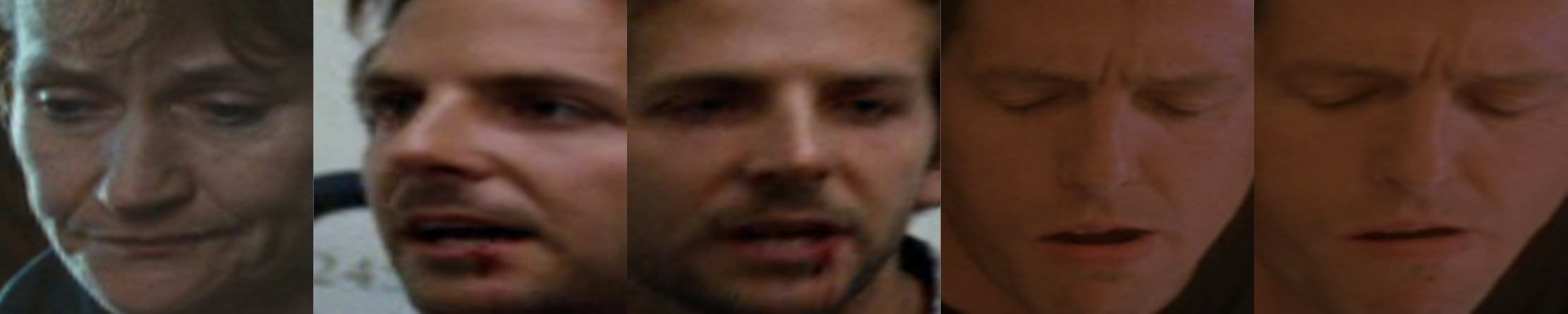}
		    \end{minipage} & Sad, Sad, Surprise, Sad, Neutral\\
		    
		     Fear& \begin{minipage}{.3\textwidth}
		   \includegraphics[width=5cm]{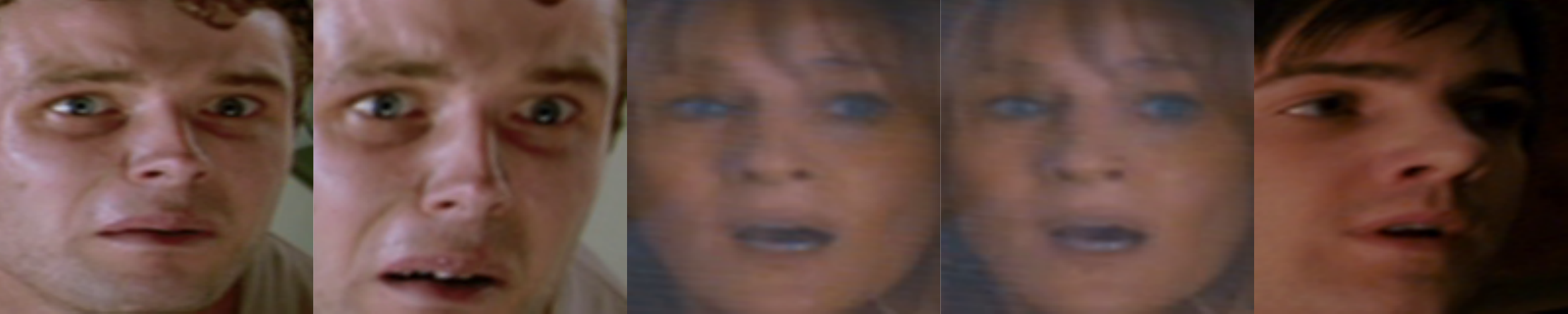} 
		   \end{minipage}
		   & \begin{minipage}{.3\textwidth}
		   \includegraphics[width=5cm]{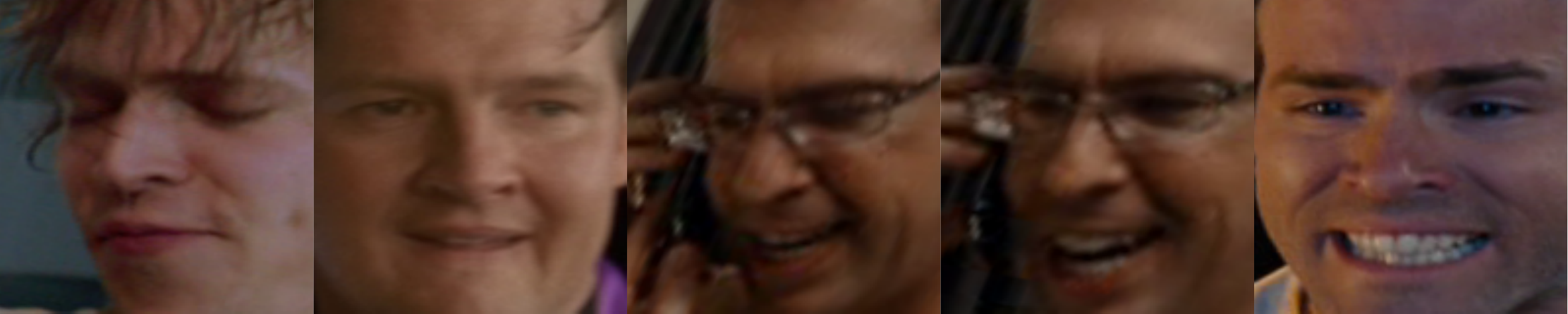}
		    \end{minipage} & Happy, Happy, Neutral, Angry, Happy\\
		     
		     Happy& \begin{minipage}{.3\textwidth}
		   \includegraphics[width=5cm]{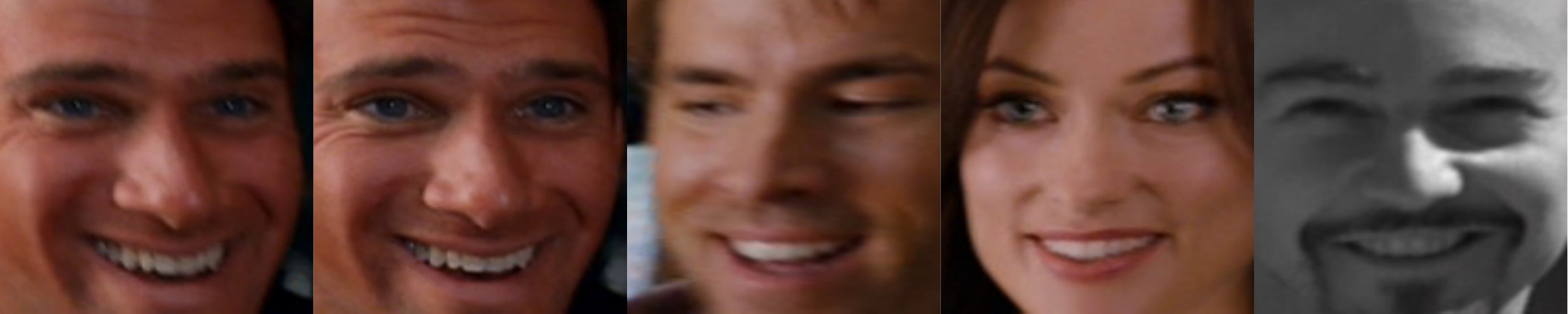} 
		   \end{minipage}
		   & \begin{minipage}{.3\textwidth}
		   \includegraphics[width=5cm]{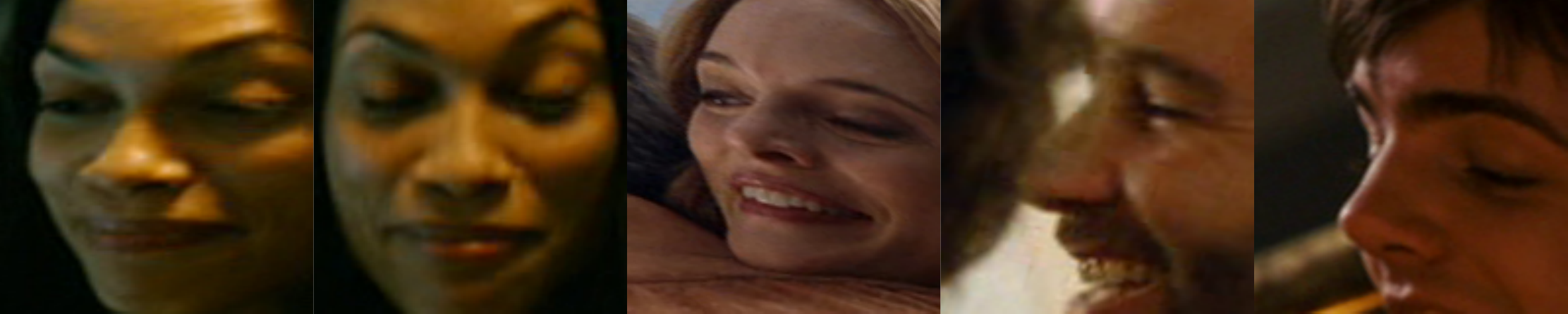}
		    \end{minipage} & Sad, Neutral, Neutral, Sad, Angry\\
		     Neutral& \begin{minipage}{.3\textwidth}
		   \includegraphics[width=5cm]{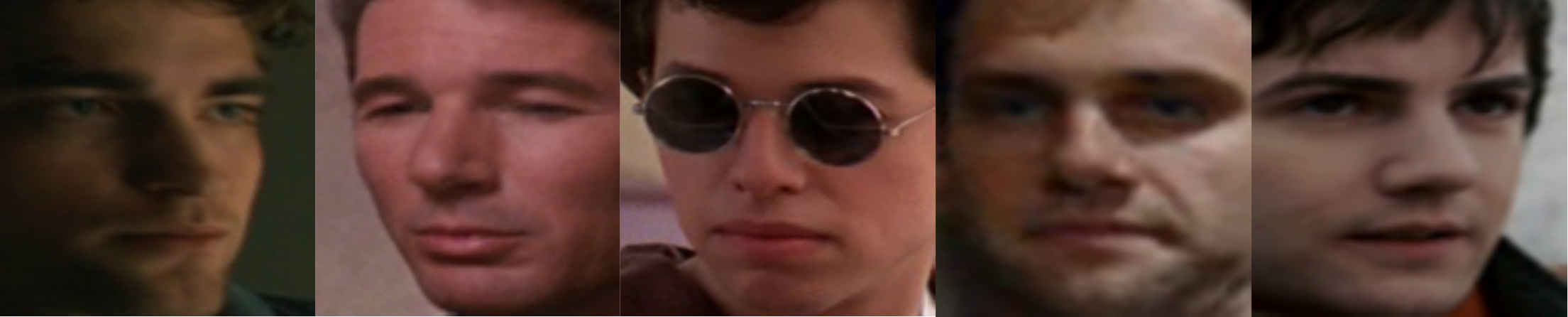} 
		   \end{minipage}
		   & \begin{minipage}{.3\textwidth}
		   \includegraphics[width=5cm]{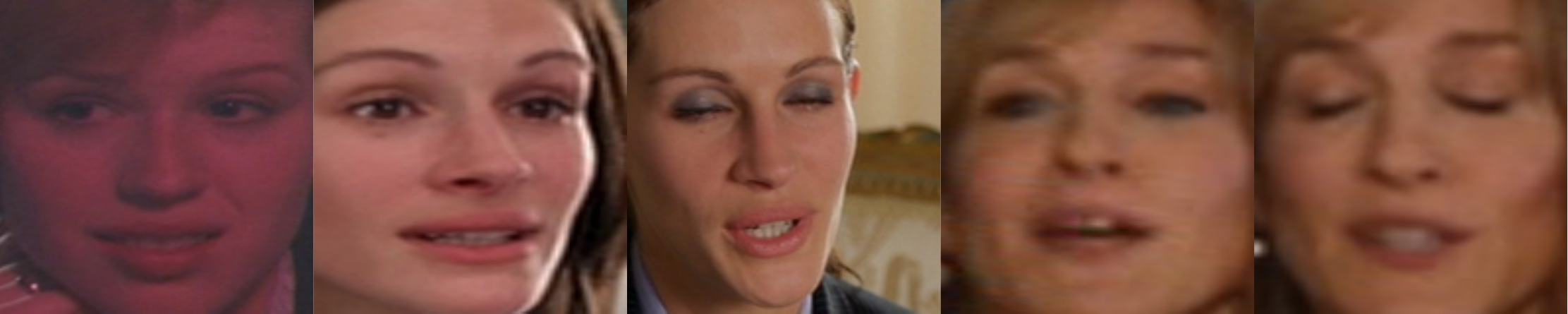}
		    \end{minipage} & Angry, Happy, Happy, Happy, Happy\\
		     Sad& \begin{minipage}{.3\textwidth}
		   \includegraphics[width=5cm]{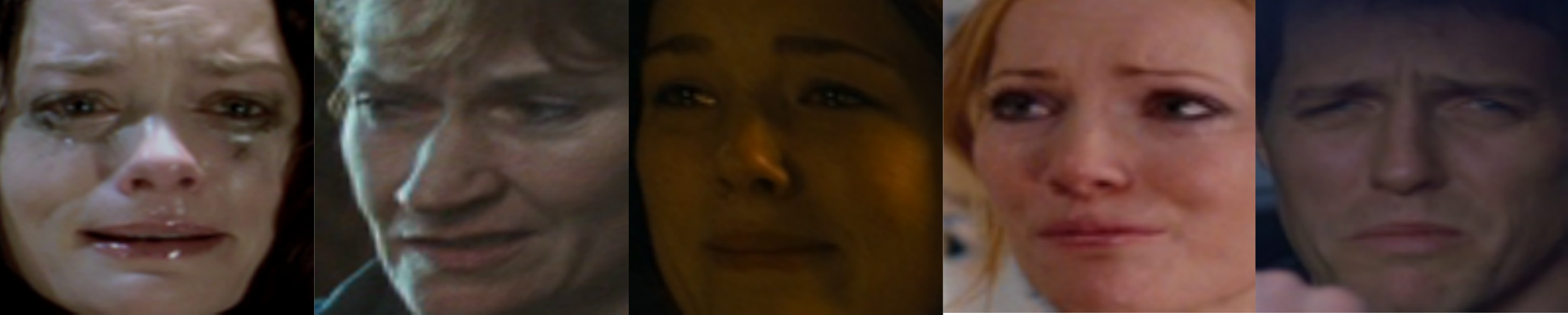} 
		   \end{minipage}
		   & \begin{minipage}{.3\textwidth}
		   \includegraphics[width=5cm]{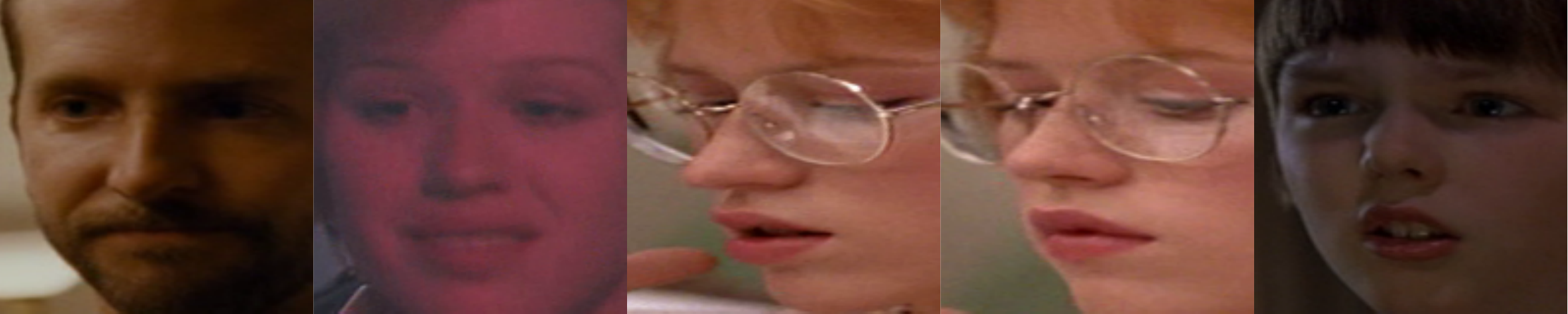}
		    \end{minipage} & Neutral, Angry, Happy, Surprise, Neutral\\
		     Surprise& \begin{minipage}{.3\textwidth}
		   \includegraphics[width=5cm]{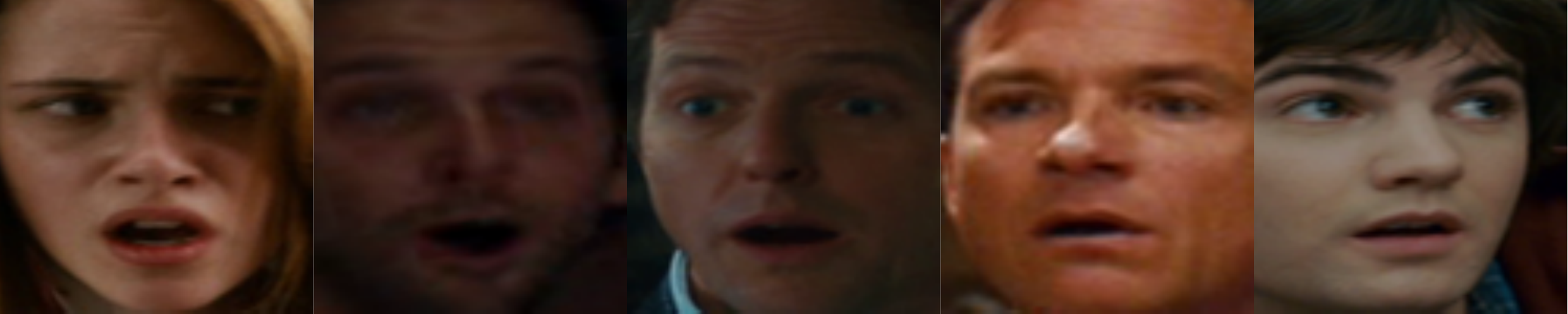} 
		   \end{minipage}
		   & \begin{minipage}{.3\textwidth}
		   \includegraphics[width=5cm]{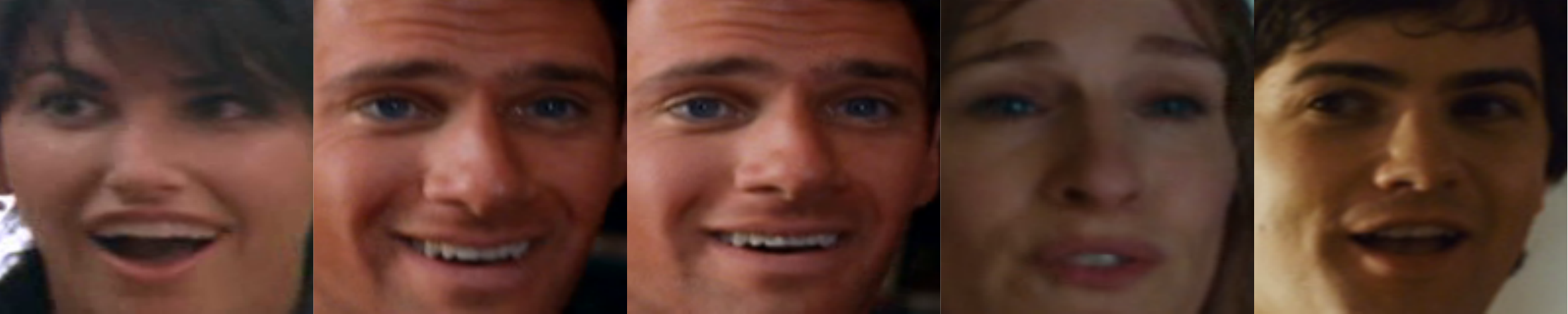}
		    \end{minipage} & Happy, Happy, Happy, Neutral, Happy\\
		    \end{tabular}
		    \caption{Samples from each class of the SFEW dataset that were most accurately and least accurately classified. The first column indicates ground truth and final column indicates predicted labels for incorrectly predicted images.}
		    \label{tab:my_label}
		\end{table*}

			\begin{figure}[ht]
					\begin{center}
						\rule{0pt}{.2in}
							\includegraphics[width=.95\linewidth,keepaspectratio]{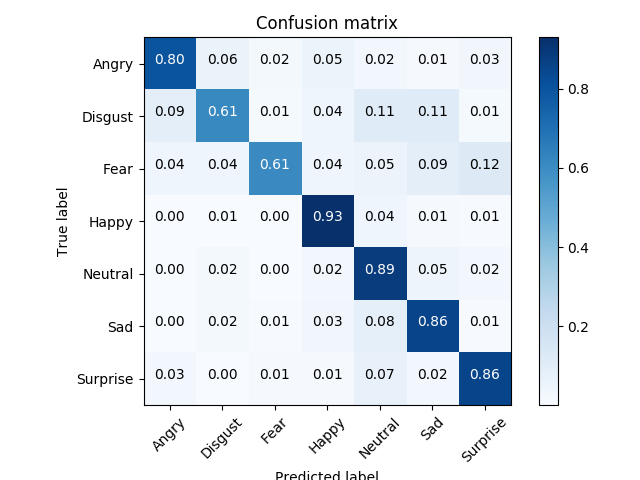}
					\end{center}
					\caption{Confusion matrix for Model-2 on the RAF dataset.}
					\label{fig:cm_raf}
				\end{figure}
				\begin{figure}
					\begin{center}
						\includegraphics[width=.95\linewidth,keepaspectratio]{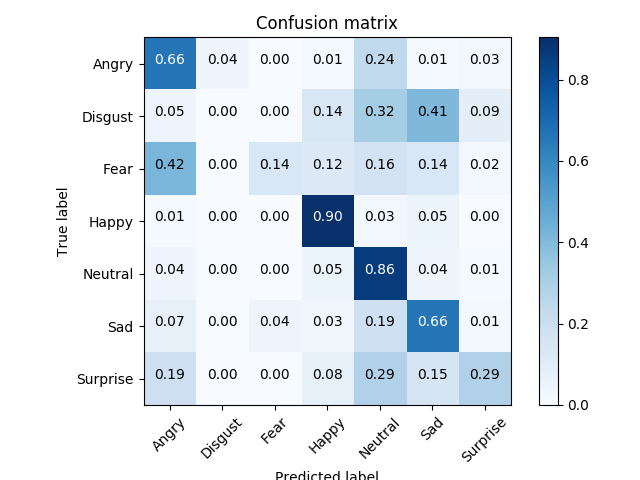}	
					\end{center}
					\caption{Confusion matrix for Model-4 on the SFEW 2.0 dataset.}
					\label{fig:cm_sfew}
				\end{figure}
		
		\subsection{Baseline Model for Video-based Recognition Problem}
		For comparing the benefits of using SPDNet over existing methods, we use kernel based PLS that used covariance matrices as features \cite{emoti14} in baseline method. 128 dimensional features were extracted from each image frame of a video and the video was modeled with a covariance matrix. Then either SPDNet or kernel based SVM with either RBF or Polynomial kernel were used for recognition. The SPDNet was able to outperform other methods.
		
		\subsection{Results on Video-based Problem}
		The results of our proposed methods, baseline method and the accuracies of other C3D and CNN-RNN models from \cite{emoti161} are presented for context. However, datasets used for those pretraining other models are not uniform, and detailed comparison of all existing methods is not within the scope of this work.
		\begin{table}[ht]
			\centering
			\begin{tabular}{lcc}
				\toprule
				Model 					& AFEW\\
				\midrule
				VGG13~\cite{emoti167}					& 57.07\\
				Single Best CNN-RNN~\cite{emoti161}		& 45.43	\\
				Single Best C3D~\cite{emoti161}			& 39.69	\\
				Single Best HoloNet~\cite{emoti1632}	& 44.57	\\
				Baseline (RBF Kernel)~\cite{emoti14}	& 45.95	\\
				Baseline (Poly Kernel)~\cite{emoti14}	& 45.43	\\
				Our proposed method (2-Bire)		    & 42.25 \\
				Our proposed method (3-Bire)		    & 44.09 \\
				Our proposed method (4-Bire)		    & 46.71 \\
				
				\midrule
				\midrule	
				Multiple CNN-RNN and C3D $^{\star\star}$~\cite{emoti161}& 51.8 \\
				VGG13+VGG16+ResNet $^{\star\star}$~\cite{emoti1632}	& 59.16 \\	
				\bottomrule
			\end{tabular}
			\caption{Video-based recognition accuracies for various single models and fusion of multiple models. Here the results of the methods marked with $^{\star\star}$ were obtained either by score level or feature level fusion of multiple models.}
			\label{table:2step}
		\end{table}
		As seen from Table \ref{table:2step}, our model was able to slightly surpass the results of the base line model. Our method also performed better than all single models that were trained on publicly available training dataset. The network from \cite{emoti167} was trained on private dataset containing an order of magnitude more samples. As a side experimentation, we also introduced covariance pooling to C3D model in \cite{emoti161} and did not obtain any improvement. We obtained accuracy of $39.78\%$.
		
% 		The results of the approaches marked with $^{\star\star}$ were obtained either by score level or feature level fusion of multiple models.

		\begin{figure}
			    \begin{center}
			    \includegraphics[width=.95\linewidth,keepaspectratio]{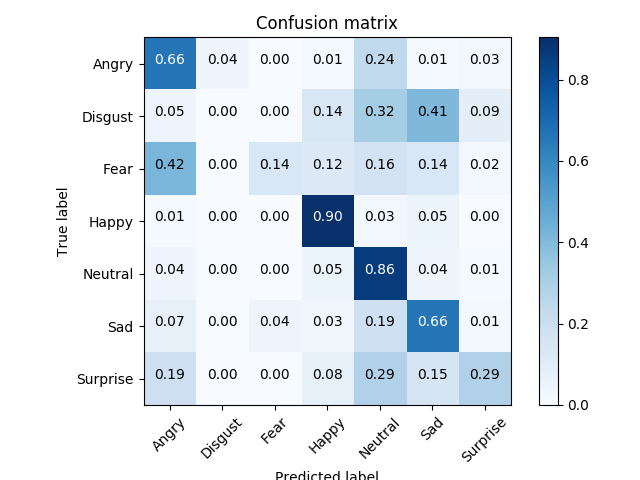}	
			    \end{center}
			\caption{Confusion matrices for our method (4-Bire) on the AFEW dataset.}
			\label{fig:cmTemporal}
		\end{figure}
		
		\section{Conclusion}

		In this work, we exploit the use of SPDNet on facial expression recognition problems. As shown above, SPDNet applied to covariance of convolutional features can classify facial expressions more efficiently. We study that second-order networks are better able to capture facial landmark distortions. Similarly, covariance matrix computed from image feature vectors were used as input to SPDNet for video-based facial expression recognition problem.
		
		We were able to obtain state-of-the-art results on image-based facial expression recognition problems on the SFEW 2.0 and RAF datasets. In video-based facial expression recognition, training SPDNet on image-based features was able to obtain results comparable to state-of-the-art results.
		
\begin{figure}[h]
			\begin{center}
				\rule{0pt}{.2in}
					\includegraphics[width=.9\linewidth,keepaspectratio]{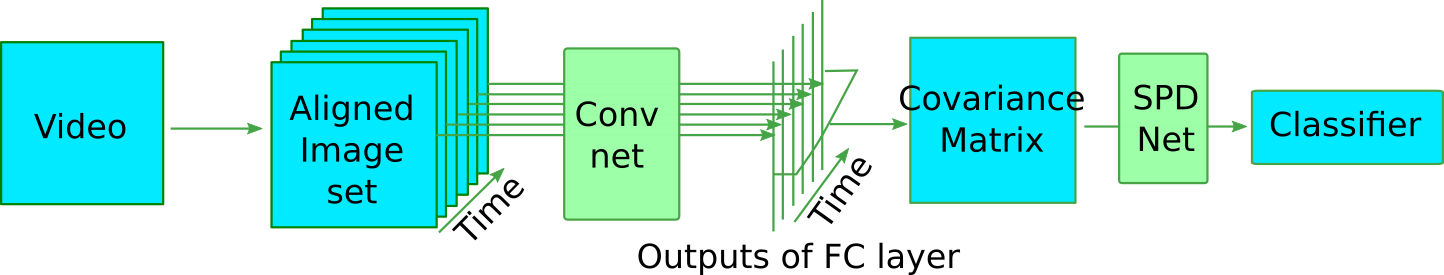}
			\end{center}
			\caption{Architecture for end-to-end training on videos directly.}
			\label{fig:e2e}
		\end{figure}

		In the context of video-based facial expression recognition problem, architecture presented in Figure~\ref{fig:e2e} can be trained in end-to-end training. Though with brief experimentation, we were able to obtain accuracy of only $32.5\%$ which is worse than the score reported ~\cite{spdnet}. It is likely to be a result of relatively small size of AFEW dataset compared to parameters in the network. Further work is necessary to see if training end-to-end using joint convolutional net and SPD net can improve results.
		
		\section{Further Works}
		
		In this work, we leveraged covariance matrix to capture second-order statistics. As studied in \cite{huang2015log}, Gaussian matrix is able to further improve the effectiveness of second-order statistics. Formally, the SPD form of Gaussian matrix can be computed by 
		\begin{equation}
		\mathbf{G}=\begin{pmatrix}
		\mathbf{\Sigma}+\mathbf{\mu}\mathbf{\mu^T} & \mathbf{\mu} \\
		\mathbf{\mu^T} & 1
		\end{pmatrix},
		\end{equation}
		where $\mathbf{\Sigma}$ is the covariance matrix defined in Eqn.~\ref{eqn:1}, and 
		\begin{equation}
		\mathbf{\mu}=\sum_{i=1}^{n}\mathbf{f_i} 
		\end{equation} is the mean of the samples $\mathbf{f_1},\mathbf{f_2},\dots,\mathbf{f_n}$ captures both first-order and second-order statistics. It was also employed in \cite{epfl} to develop second-order convolutional neural networks. Extending current work from covariance pooling to Gaussian pooling would be an interesting direction and should theoretically improve results.

{\small
\bibliographystyle{ieee}
\bibliography{egbib}
}

\end{document}